\documentclass{article}

\usepackage{arxiv}

\usepackage[utf8]{inputenc} 
\usepackage[T1]{fontenc}    
\usepackage{hyperref}       
\usepackage{url}            
\usepackage{booktabs}       
\usepackage{amsfonts}       
\usepackage{nicefrac}       
\usepackage{microtype}      
\usepackage{lipsum}		
\usepackage{graphicx}
\usepackage{natbib}
\usepackage{doi}
\usepackage{amsmath}
\usepackage{multirow}
\usepackage{soul}
\usepackage{placeins}

\title{FD-LLM: Large Language Model for Fault Diagnosis of Machines}

\author{\href{https://orcid.org/0000-0002-0095-9401}{\includegraphics[scale=0.06]{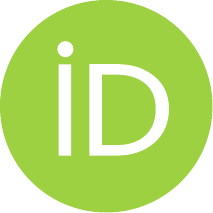}\hspace{1mm}HAMZAH A.A.M. QAID} \\
	School of Mechatronic Engineering\\
	China University of Mining and Technology\\
	Xuzhou, China \\
	\texttt{fb23050001e@cumt.edu.cn} \\
	\And
	\href{https://orcid.org/0000-0003-1129-1109}{\includegraphics[scale=0.06]{orcid.pdf}\hspace{1mm}Bo Zhang}\\
	School of Computer Science and Technology\\
	China University of Mining and Technology\\
	Xuzhou, China \\
	\texttt{zbcumt@cumt.edu.cn} \\
   \And
	\href{https://orcid.org/0000-0002-3787-1673}{\includegraphics[scale=0.06]{orcid.pdf}\hspace{1mm}Dan Li} \\
	School of Software Engineering\\
	Sun Yat-Sen University\\
	Zhuhai, China \\
	\texttt{lidan263@mail.sysu.edu.cn} \\
    \And
	See-Kiong Ng \\
	Institute of Data Science\\
	National University of Singapore\\
	Singapore, Singapore \\
	\texttt{seekiong@nus.edu.sg} \\   
    \And
	Wei Li \\
	School of Mechatronic Engineering\\
    China University of Mining and Technology\\
	Xuzhou, China \\
	\texttt{liwei\_cmee@163.com} \\
}

\begin{document}
\maketitle

\begin{abstract}
Large language models (LLMs) are effective at capturing complex, valuable conceptual representations from textual data for a wide range of real-world applications.  However,  in fields like Intelligent Fault Diagnosis (IFD), incorporating additional sensor data---such as vibration signals, temperature readings, and operational metrics---is essential but it is challenging to capture such sensor data information within traditional text corpora.  This study introduces a novel IFD approach by effectively adapting LLMs to numerical data inputs for identifying various machine faults from time-series sensor data.  We propose FD-LLM, an LLM framework specifically designed for fault diagnosis by formulating the training of the LLM as a multi-class classification problem. We explore two methods for encoding vibration signals: the first method uses a string-based tokenization technique to encode vibration signals into text representations, while the second extracts statistical features from both the time and frequency domains as statistical summaries of each signal.  We assess the fault diagnosis capabilities of four open-sourced LLMs based on the FD-LLM framework, and evaluate the models’ adaptability and generalizability under various operational conditions and machine components, namely for traditional fault diagnosis, cross-operational conditions, and cross-machine component settings. Our results show that LLMs such as Llama3 and Llama3-instruct  demonstrate strong fault detection capabilities and significant adaptability across different operational conditions, outperforming state-of-the-art deep learning (DL) approaches in many cases.	    
\end{abstract}


\section{Introduction}
 Driven by automation and advanced technologies to maximize productivity and efficiency, modern industrial systems have evolved into highly sophisticated networks with growing complexity that amplifies the risks associated with machine faults.  Even minor faults may potentially lead to significant downtime, financial losses, and safety hazards.  Timely and accurate detection of potential issues is crucial to maintain the integrity and long-term viability of industrial operations in this increasingly competitive and complex landscape.

 Over the past two decades, numerous studies have focused on developing Intelligent Fault Diagnosis (IFD) systems using traditional machine learning (ML) and advanced deep learning (DL) algorithms. With their effectiveness in processing time-series data and enhancing the automation of fault detection \citep{Liu2018ArtificialIF,zhao2020deep}, these systems are now widely used in industry. 
 However, ML and DL algorithms come with several limitations. They can produce uncertain results, struggle to handle complex and heterogeneous data sources, and often fail to provide maintenance engineers with actionable insights into the root causes of faults. Additionally, ML and DL models tend to have limited generalization ability across different operational conditions or machines, often requiring extensive retraining or fine-tuning when applied to new equipment or changing environments. These challenges can hinder their effectiveness in critical industrial contexts, where rapid adaptability and high interpretability are essential.

Recently, large language models (LLMs) \citep{Zhao2023ASO,Zhou2023ACS}, such as GPT-2 \citep{Radford2019LanguageMA}, Llama-2 \citep{Touvron2023Llama2O}, and Qwen-2 \citep{Yang2024Qwen2TR}, have achieved groundbreaking advances in the field of natural language processing (NLP). With their exceptional language comprehension and near-human conversational abilities, LLMs have shown great promise in advancing toward General Artificial Intelligence (GAI) in complex, data-intensive environments. Although originally designed to process large volumes of unstructured data like text and images, the advancements in LLMs also offer a promising alternative for intelligent fault diagnosis, through the use of encoding methods such as  {\it string-based tokenization} \citep{Gruver2023LargeLM}, which transforms time-series data into numerical strings that LLMs can interpret as natural language inputs, {\it modality-specific encoding\/}, which embeds non-text data modalities including time-series into LLMs’ token space using pre-trained encoder \citep{Belyaeva2023MultimodalLF} or neural network \citep{sun2023test}, and {\it statistical summarization\/} \citep{jin2023time}, which generates statistical summaries of time-series data and serializes them as text.

Several time series LLMs have recently been developed using these encoding methods.  They generally fall into two categories: general-purpose and domain-specific applications. General-purpose models aim to tackle a wide range of tasks, including forecasting \citep{ansari2024chronos,jin2023time}, anomaly detection \citep{zhou2023one}, and classification \citep{sun2023test} for various application domains, while domain-specific LLMs are designed for specialized applications in specific domains, such as Health-llm \citep{Kim2024HealthLLMLL} for health prediction tasks, and StockGPT \citep{Mai2024StockGPTAG} and Stock-Chain \citep{Li2024AlphaFinBF} for financial analysis tasks such as stock trend prediction. 

As research continues to uncover the broader capabilities of LLMs,  expanding their use to more complex challenges---such as intelligent fault diagnosis that involves the integration of time-series data---remains a key area for future exploration that can bring about great impact in many other domains.  For example, the strong zero-shot generalizability in instruction-tuned LLMs is crucial for fault diagnosis, where there are limited data yet multiple diverse operational conditions and machine components.  Unlike the DL models which would require much data for additional training or fine-tuning, if LLMs can be leveraged to create reliable fault predictors that perform effectively across different operational conditions and machine components without extensive retraining, it would greatly improve diagnostic efficiency.  
As such, this paper investigates the potential of LLMs in fault diagnosis in various operational conditions and machine component settings, with the following  contributions: 

(1) We present FD-LLM, a framework designed to enable large language models (LLMs) to adapt to fault diagnosis tasks. This new approach is formulated as a multi-class classification problem, where we fine-tune LLMs using vibration signals and instruction prompts to identify potential faults; 

(2) We incorporate two representation methods for vibration signals into  FD-LLM. The first method uses a string-based tokenization technique, converting FFT-processed vibration data into text representations that are compatible with LLM input. The second method extracts statistical summaries from both the time and frequency domains of vibration signals, resulting in a text paragraph that can be processed by the  LLMs;

(3) We conduct an extensive evaluation of several leading open-source LLMs,  including Llama3-8B, Llama3-8B-instruct, Qwen1.5-7B, and Mistral-7B-v0.2. We assess the models’ performance in fault diagnosis under three settings: traditional fault diagnosis settings followed in DL and ML models; cross-dataset settings to assess the models’ generalizability across various operational conditions and machine components settings.

\section{Related work}
\textbf{Intelligent Fault Diagnosis (IFD).} IFD has been a critical area of research in industrial maintenance, with significant advancements driven by the application of ML and DL-based techniques. Traditional ML-based fault diagnosis methods typically rely on handcrafted features derived from sensor data through signal processing techniques \citep{Wang2017MatchingSW} and fault identification \citep{Sun2018SparseDS} using ML models. Common models used in these approaches include support vector machine (SVM) \citep{Wu2006CompoundRM,Tang2010MultifaultCB}, K-Nearest neighbour (k-NN) \citep{Wang2016KnearestNB,Pandya2013FaultDO}, Naïve Bayes classifier\citep{Zhao2009DiagnosisFT,Muralidharan2012ACS}, and artificial neural networks (ANN) \citep{Mrugalski2008ConfidenceEO,Rafiee2007INTELLIGENTCM}. While these techniques have shown good performance, they are constrained by the quality of feature extraction and the challenges of managing large, complex datasets.

The subsequent advent of DL which enabled the automatic extraction of features directly from raw sensor data has led to a paradigm shift in IDF.
Convolutional neural networks (CNNs), for instance, have been effectively employed to detect and classify faults by learning spatial hierarchies of features from vibration signals \citep{Zhang2017AND,Abdeljaber2017RealtimeVS,Ince2016RealTimeMF}. Recurrent neural networks (RNNs) were used to capture temporal dependencies in the time-series data \citep{Yuan2016FaultDA,Zhao2016MachineHM}. Despite their success, DL models typically require large amounts of labelled data for training, and they may struggle to generalize across different machines or operational conditions without extensive retraining. Furthermore, as  "black boxes," they provide little interpretability.

Recently, domain adaptation, which aims to improve the performance of DL models when applied to new, unseen domains by transferring knowledge from a source domain (where labelled data is abundant) to a target domain (where labelled data is scarce), has gained increasing attention as a promising method to address the adaptability challenges of DL models in IFD \citep{Zhao2019ApplicationsOU}. Several studies have explored domain adaptation in various experimental settings, including Closed-Set Domain Adaptation (CSDA) \citep{Zhang2022SupervisedCL}, Partial Domain Adaptation (PDA) \citep{Wang2022ABA}, and Open Set Domain Adaptation (OSDA) \citep{Guo2022AnOF}. While these domain adaptation techniques have shown promise in enhancing the generalization capabilities of DL models, they still face several challenges. While these techniques have shown promise in improving the generalization capabilities of DL models, they still face several challenges. For instance, the success of domain adaptation models often relies heavily on the similarity between the source and target domains; significant differences between them can lead to suboptimal performance. Furthermore, these models may still require retraining or fine-tuning, which can be time-consuming and resource-intensive. Ensuring consistent and stable adaptation of the DL models across varying fault conditions and domains remains a critical challenge for their practical deployment. 

\textbf{Time Series Data with LLMs.} Recently, it has been demonstrated that large language models (LLMs) can be applied to time series or tabular data modeling using techniques such as direct prompting and multimodal fine-tuning \citep{jin2024position}. Direct prompting 
involve preprocessing non-textual data into representations that fit the token space of LLMs, incorporating these representations into prompt templates to create the final input, and then feeding the processed input into the LLM to generate responses.  
On the other hand, multimodal fine-tuning techniques integrate the capabilities of LLMs’ text processing with time-series modality, where the text modality serves as task instructions or prompts that describe the time-series modality guiding LLMs to learn representation from the input. Multimodal fine-tuning follows three steps, typically including a pre-processing step where numerical signals are patched and tokenized, followed by a fine-tuning step tailored for general time series tasks or domain-specific applications, and post-processing step responsible for inference of the prediction results. 

In both techniques, the pre-processing step aims to bridge the modality gap between numerical or sensory data and LLMs input. Spathis et al.\citep{spathis2024first} addressed the modality gap between text and numerical data by employing lightweight embedding layers and prompt design.  Gruver et al.\citep{Gruver2023LargeLM} proposed a highly effective yet simple string-based tokenization method, converting numerical time-series values into text-like representations. 
Ansari et al.\citep{ansari2024chronos} encoded time series into fixed vocabulary using a sequence of reversible steps including scaling and quantization, then trained transformer-based models through cross-entropy loss, achieving strong zero-shot forecasting performance.  

Another line of studies introduces modality-specific encoding by utilizing lightweight adaptation layers. For example, Time-LLM\citep{jin2023time} reprogrammed time-series data into the LLMs language space and used text prompts as prefixes, improving LLM performance in forecasting tasks. Likewise, TEST\citep{sun2023test} resolves embedding inconsistencies by developing a time-series encoder, leveraging alignment contrasts with soft prompts for efficient fine-tuning of frozen LLMs. However, employing modality-specific encoding raises a number of possible issues, including the complexity of multimodal fine-tuning frameworks, the computational cost, particularly when handling long time-series signals, and the possibility of information imbalance, where some modalities may be underrepresented or dominate the model's attention.

\textbf{Our work.} Our FD-LLM framework is specifically designed for fault diagnosis, through framing the training (i.e., fine-tuning) of existing open-source LLMs as a multi-class classification problem.
FD-LLM incorporates two methods for pre-processing vibration signals. In the first method, we apply the Fast Fourier Transform (FFT) to the vibration signals and calculate the magnitudes from the FFT results, generating feature vectors with non-negative values. These vectors are then converted into string-based representations, following the approach outlined by \citep{Gruver2023LargeLM}. In the second method, we extract various metrics, including statistical features from the time domain and spectral features from the frequency domain (collectively referred to as statistical features). These features are then summarized in a tabular-like format. Inspired by \citep{Belyaeva2023MultimodalLF}, we serialize each feature attribute along with its corresponding feature name into a prompt template that is compatible with LLM input. Through utilizing these two encoding methods, the FD-LLM framework treats vibration signals as text modality inputs, thereby avoiding the complexity and embedding inconsistencies or pseudo-alignment behavior introduced by modality-specific encoding methods.

\section{Methods}
\subsection{Problem definition}
We formulate the fine-tuning of large language models (LLMs)  as a multi-class classification problem. The objective is to predict the fault type \( t \in T \), where \( T = \{t_1, t_2, \ldots, t_k\} \) represents the set of possible fault categories, given a set of fault samples \( X = \{x_1, x_2, \ldots, x_n\} \), and their corresponding prompts \( P = \{p_1, p_2, \ldots, p_n\} \).

In other words, the task is to train a fault prediction system \( \text{llm} : (X, P) \rightarrow T \), where the prediction system \( \text{llm} \) maps each pair of fault sample \( x_i \) and prompt \( p_i \) to a predicted fault label \( \hat{t_i} \). This system is to generate a prediction that corresponds to one of the predefined fault categories, expressed mathematically as follows:

\begin{equation}
    \hat{t_i} = \text{llm}(I = g(x_i, p_i)), \quad \text{for } i = 1, 2, \ldots, n
    \label{eq:fd-llm}
\end{equation}

\noindent where \( g(\cdot) \) is a function that integrates an input sample \( x_i \) and its corresponding prompt \( p_i \), generating the final input \( I \). Our objective is to leverage both the raw fault data and the additional context provided by the prompts to assess LLMs’ potential in fault diagnosis and investigate whether leveraging machine specification enhances the performance across different work conditions and machine components.

\subsection{FD-LLM framework}
\begin{figure}[t!]
    \centering
    \includegraphics[width=\linewidth]{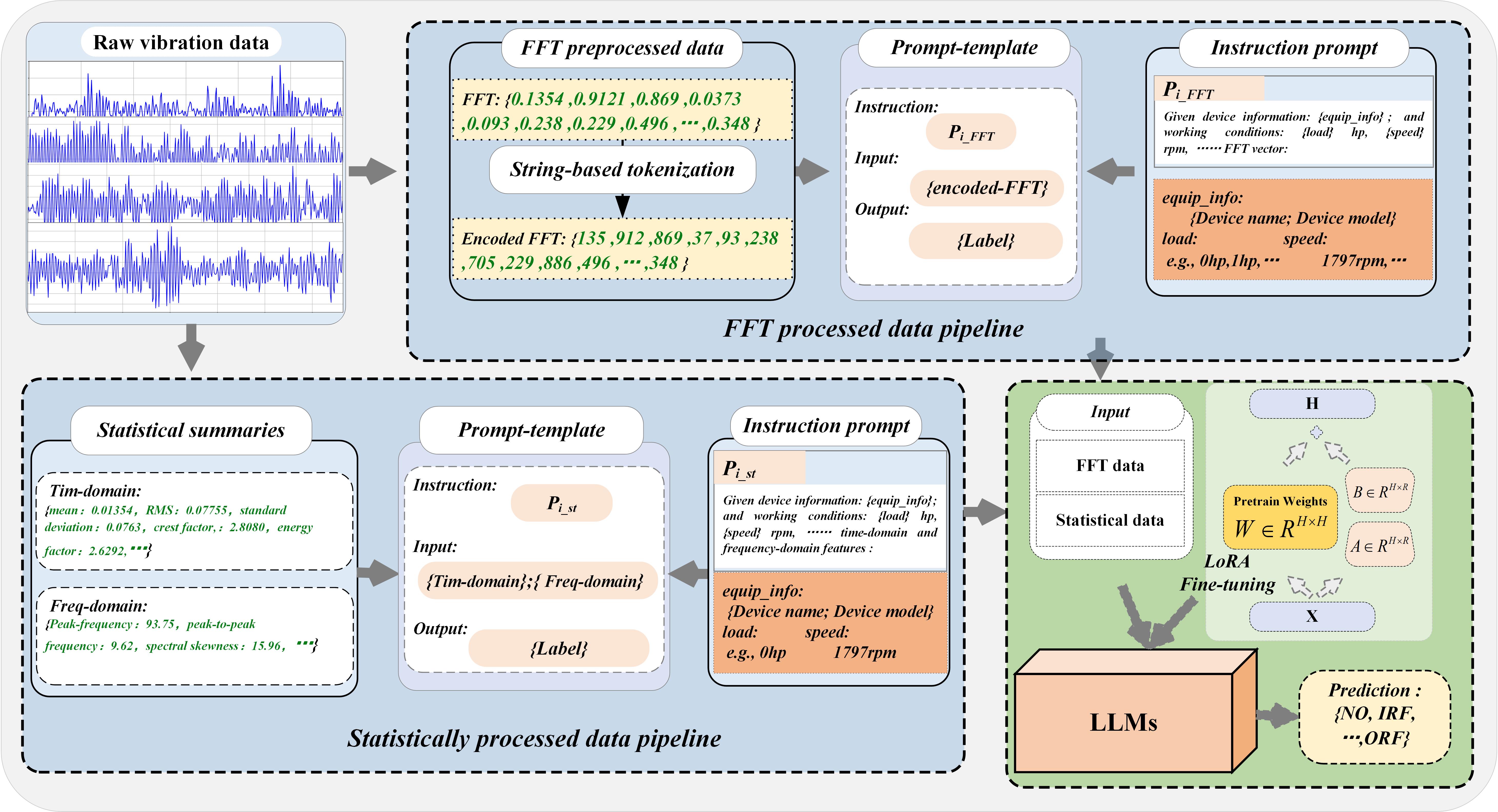} 
    \caption{An illustration of the FD-LLM framework using both FFT-processed and statistically processed data pipelines.}
    \label{fig: FD_LLM}
\end{figure}

 Figure~\ref{fig: FD_LLM} illustrates our proposed pipeline for FD-LLM (Fault Diagnosis Large Language Model), designed to assess the health state of mechanical equipment by predicting potential faults based on vibration signals. The process begins with the preprocessing of vibration signals into representative samples, either by applying the FFT or by generating statistical summaries from both the time and frequency domains, thereby preparing the signals for LLM input. These processed samples are then combined with carefully crafted instruction prompts to create the final input for the LLM to analyze and generate predictions for the potential fault types. Specifically, for an input sample $x_i$, which could either be an FFT vector or a row of statistical features, along with the corresponding prompt $P_{i\_fft}$ or $P_{i\_st}$, the LLM is fine-tuned to analyze the data and output fault type predictions. 
 
 Note that the prompt
 contains essential contextual information, including equipment specifications (e.g., name, model, geometric parameters) and the machine’s operating conditions, such as speed and load. By embedding this contextual information alongside each signal sample in the input prompt, the LLM can utilize both the raw data and the relevant operational details to make accurate fault predictions.  Moreover, incorporating such information leads to more robust and reliable diagnostic results by the LLM across various machine components operating under diverse conditions, as it enables the LLM  to consider how different operational conditions and machine component types affect performance.
\subsubsection{Data pre-processing}
\label{set:3.2.1}
As mentioned, FD-LLM utilizes two methods for pre-processing vibration signals to generate proper time-series representations that are LLM-ready. We discuss these two methods in detail as follows.

\textbf{(1)	FFT pre-processing.} 
The raw vibration signals are first subjected to random sampling, segmenting each signal into multiple overlapping or non-overlapping windows, each of which is then analyzed to capture its frequency characteristics. The objectives are: (i) to reduce the overall length of the vibration signals, ensuring they do not exceed the LLMs' maximum sequence length (e.g., MISTRAL has a maximum context length of 32k tokens, and Llama3 supports up to 8k tokens) while augmenting the training samples; and (ii) to produce a normalized frequency representation for each segment, which can then be incorporated into the prompt template after a proper encoding.  

Given a set of time-domain signals \( \{ x^{(j)}(n) \}_{j=1}^{J} \), where \( J \) is the total number of signals, we divide each signal into \( K \) segments, with each segment having a length of \( L \) data points. For instance, if the total length of the \( j \)-th signal is \( N^{(j)} \), we can denote the \( k \)-th segment of the \( j \)-th signal as \( x_k^{(j)}(n) \), where \( n = 0, 1, \ldots, L-1 \). For each segment \( x_k^{(j)}(n) \), we compute the FFT, which is essentially an efficient implementation of the Discrete Fourier Transform (DFT). The DFT can be expressed mathematically as:
\begin{equation}
X_k^{j}[m] = \sum_{n=0}^{L-1} x_k^{(j)}(n) \cdot e^{-i \frac{2\pi}{L} m n}, \quad m = 0, 1, \ldots, L-1
\label{eq:dft_samples}
\end{equation}
where \( i \) is the imaginary unit. 

We further compute the magnitude \( P_k^{(j)}[m]\) for each FFT coefficient as \( P_k^{(j)}[m] = |X_k^{(j)}[m]| \), representing the amplitude of each frequency in the segment. This results in FFT-processed samples that include only positive values, thereby avoiding the need for unnecessary sign tokenization. To ensure consistency across different segments, we scale the magnitudes by dividing by the segment length \( L \): \( Y_k^{(j)}[m] = \frac{P_k^{(j)}[m]}{L} \). For each segment \( k \) of the \( j \)-th signal, the output \( Y_k^{(j)}[m]\) comprises \( L \) normalized values, one for each frequency component \( m \). Finally, a set of FFT-processed samples for all signals is obtained and represented as follows:: \(\mathcal{Y} = \{ \mathcal{Y}^{(j)} \}_{j=1}^{J},\quad \text{where } \mathcal{Y}^{(j)} = \{ Y_k^{(j)} \}_{k=1}^{K}, \text{ for } j = 1, 2, \ldots, J,\) and \(Y_k^{(j)} = [Y_k^{(j)}[0], Y_k^{(j)}[1], \ldots, Y_k^{(j)}[L-1]]\).

\begin{figure}[t!]
    \centering
    \includegraphics[width=\linewidth]{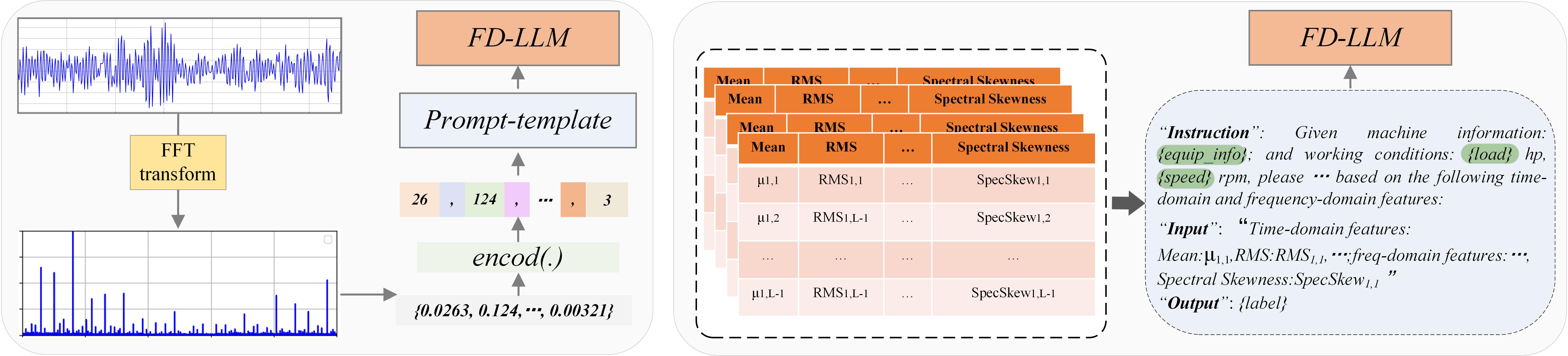} 
    \caption{\textbf{Left:} Illustrates the FFT pre-processing steps. For illustration purposes, we use a complete vibration signal and its corresponding FFT transformation from an outer race fault of the bearings. We select to use a length of 512 for each segment, and the number of decimal places \(D\) is set to 3. \textbf{Right:} Presents the statistically processed data, where each table contains the statistical features of segments from a certain signal collected under certain machine health state. Each feature name and its corresponding value are serialized into the input element of the prompt template for every segment in the tables.}
    \label{fig: data-pre}
\end{figure}

The FFT-processed samples are subsequently encoded and incorporated into a structured prompt template to include essential machine information, operating conditions, encoded FFT samples, and the corresponding label. The prompt template consists of three main elements: "\textit{instruction}", "\textit{input}", and "\textit{output}", as displayed in Table \ref{table:1}.  Please also refer to Figure \ref{fig: data-pre} for a visual illustration of the FFT pre-processing steps using a complete vibration signal. 

     \textbf{Instruction}: A prompt \(p_{i\_fft}\) that integrates an instruction or task query along with the machine specifications \(\{\text{equip-info}\}\), operating conditions (workload \(\{\text{load}\} \, \text{hp}\) and rotation speed \(\{\text{speed}\} \, \text{rpm}\)) into a cohesive paragraph using the function \( f(\cdot) \) in formula \ref{eq:2}.
    
\textbf{Input:} This element, represented as \(x_i\), consists of the FFT samples, which are transformed into a suitable format for integration into the LLM input through a string encoding function  \( \texttt{encode}(\cdot) \), as expressed in formula \ref{eq:3}.

\textbf{Output:} Captures the corresponding label for each pair of the previous two elements.

\begin{table}[h!]
\centering
\caption{Instruction prompts used to generate the training data}
\label{table:1}
\begin{tabular}{p{5cm} p{10cm}}
\toprule 
\textbf{Dataset} & \textbf{Prompt-template} \\ 
\cmidrule(lr){1-2} 
\vspace{0.2cm}FFT processed samples & 
\texttt{"Instruction"}: Given machine information: \{equip\_info\}; and working conditions: \{load\} hp, \{speed\} rpm, please predict the operating status of the bearing based on the following FFT vector. \newline
\texttt{"Input"}: \{FFT\} \newline \texttt{"Output"}: \{label\}\\ 
\cmidrule(lr){1-2} 
\vspace{0.2cm}Statistically processed samples & 
\texttt{"Instruction"}: Given machine information: \{equip\_info\}; and working conditions: \{load\} hp, \{speed\} rpm, please predict the operating status of the bearing based on the following time-domain and frequency-domain features. \newline
\texttt{"Input"}: \{Tim-domain features\}; \{freq-domain features\} \newline
\texttt{"Output"}: \{label\} \\ 
\bottomrule 
\end{tabular}
\end{table}
\vspace{-8pt}
\begin{align}
p_{i\_fft} &= f(\text{equip\_info}, \text{load}, \text{speed}, q_{_{fft}}) \label{eq:2} \\
x_i^{(j)} &= \text{encode}(Y_k^{(j)}[0], Y_k^{(j)}[1], \ldots, Y_k^{(j)}[L-1]) \label{eq:3}\\
I &= g(x_i^{(j)}, p_{i\_fft}) \label{eq:4}
\end{align}
\noindent where \(f(\cdot)\) is the function that aggregates the equipment details \text{equip\_info}, \text{load}, and \text{speed},  and task query \(q\). \(x_i^{(j)} \)is the encoded FFT vector. \(g(\cdot)\) concatenates the instruction prompt \(p_i^{(j)} \) and encodes FFT vector into a single input \(I \) . In Section \ref{sec:st}, we introduce a method for string encoding of FFT samples.

(2) \textbf{Statistical pre-processing. } We extract a set of 15 statistical features: 10 from the time domain and 5 from the frequency domain. The time-domain features include the mean, root mean squared (RMS), standard deviation, crest factor, skewness, shape factor, kurtosis, peak-to-peak value, energy factor, and impulse factor. The frequency-domain features include peak frequency, peak-to-peak frequency, spectral kurtosis, spectral bandwidth, and spectral skewness.

For each segment \( x_k^{(j)}(n) \),  we derive a total of 15 features, resulting in a feature vector \( \mathbf{F}_{k}^{(j)} \) that encapsulates both time-domain and frequency-domain statistics:

\begin{equation}
\mathbf{F}_{k}^{(j)} = 
\left[ 
\mathbf{F}_{k,\text{time}}^{(j)}, 
\mathbf{F}_{k,\text{freq}}^{(j)} 
\right]
\label{eq:st}
\end{equation}

where
\begin{equation}
\mathbf{F}_{k,\text{time}}^{(j)} = 
\left[ 
\mu_{k}^{(j)}, 
\text{RMS}_{k}^{(j)}, 
\sigma_{k}^{(j)}, 
\text{CF}_{k}^{(j)}, 
\text{Skew}_{k}^{(j)}, 
\text{SF}_{k}^{(j)}, 
\text{Kurt}_{k}^{(j)}, 
\text{P2P}_{k}^{(j)}, 
\text{EF}_{k}^{(j)}, 
\text{IF}_{k}^{(j)} 
\right]
\label{eq:tim_features}
\end{equation}
are the 10 time-domain features, and

\begin{equation}
\mathbf{F}_{k,\text{freq}}^{(j)} = 
\left[ 
\text{PeakFreq}_{k}^{(j)}, 
\text{P2PFreq}_{k}^{(j)}, 
\text{SpecKurt}_{k}^{(j)}, 
\text{SpecBW}_{k}^{(j)}, 
\text{SpecSkew}_{k}^{(j)} 
\right]
\label{eq:freq-domain}
\end{equation}
are the 5 frequency-domain features. The feature vectors from all segments are systematically arranged in a tabular format, with the first row listing the feature names to define each dimension of the data. Each subsequent row corresponds to an individual segment, encapsulating its extracted features as a structured vector. 

To ensure compatibility with LLMs, the feature vectors are serialized by converting them into concise textual summaries within a predefined prompt template presented in Table \ref{table:1}. Similar to the FFT-processed data, the prompt template also consists of the same three elements, where \(p_{i\_st}\)  includes a task query indicating the input being statistical summaries, and the input element  \(x_{i}\)  contains the textual summaries of the statistical feature vectors. This transformation preserves the integrity of the original data while enabling efficient input to LLMs. A visual demonstration of this process is illustrated in Figure \ref{fig: data-pre}. A detailed breakdown of the 15 features along with their respective mathematical formulas can be found in Appendix \ref{sec:app-A} Table \ref{tab:summary_features}. 

\subsubsection{Time series data encoding}
\label{sec:st}
Each LLM uses a tokenizer to convert input text into a sequence of tokens, a critical process since even small discrepancies can lead to significant changes in model behavior. One common tokenization technique is Byte-Pair Encoding (BPE), which processes input data as bit strings and generates tokens based on their frequency in the training data. However, BPE can sometimes split a single number into multiple tokens that do not align with its individual digits, which can hinder the model’s ability to interpret and perform operations on numerical data. As a result, properly encoding time series data into a textual format that ensures accurate tokenization is a crucial step for enabling LLMs to make reliable predictions.

Inspired by the method presented in \citep{Gruver2023LargeLM}, we transform the FFT-processed samples into a sequence of values that can be accurately tokenized. Given a segment 
 \( Y_k^{(j)} = [Y_k^{(j)}(0), Y_k^{(j)}(1), \ldots, \allowbreak Y_k^{(j)}(L - 1)] \), for simplicity, we use \( Y = (y_0, y_1, \ldots, y_{L-1}) \) instead. We define an encoding function \( \text{encode}(\cdot) \) that performs a series of reversible steps: sign handling, quantization, and string tokenization, as described below. 

\textbf{Sign Handling:} Although FFT magnitudes \( y_i \) are inherently non-negative, we incorporate a mechanism for handling potential sign adjustments for completeness. The sign of each spectral value is processed using conditional logic:

   \begin{equation}
       \text{sign}(y_i) = \begin{cases} 
   1, & \text{if } y_i \geq 0 \\
   -1, & \text{if } y_i < 0
   \end{cases}
   \label{eq:sign}
   \end{equation}
   
   This step ensures compatibility for scenarios where sign adjustments might be applicable. Given that FFT magnitudes are generally positive, this prevents unnecessary sign tokenization and reduces the length of the input samples.

\textbf{Quantization:} To address the potentially high precision of FFT magnitudes, we apply a quantization function \( \text{quantize}(Y, D) \) over the FFT-processed sample \(Y = (y_0, y_1, \ldots, y_{L-1}) \) to enhance computational efficiency. The parameter \(D\) specifies the number of decimal places to retain before truncation. Each magnitude \( y_i \) in the \(Y\) is truncated to \(D\) decimal places and converted into an integer representation to minimize token usage associated with floating-point values. Consequently, the quantized FFT-processed samples become \(Y_q = (y_{q0}, y_{q1}, \ldots, y_{q(L-1)})\).

\textbf{String Tokenization:} The quantized samples are then transformed into a formatted string representation with a specific separator. Given that we utilize open-source LLMs such as Llama-3, Llama3-instruct, and Mistral, which effectively manage the tokenization of numbers, there is no need to insert spaces between digits. However, handling missing values is also considered in this transformation, where a placeholder string (e.g., "NaN") replaces any missing values. Finally, we obtain an encoded representation \( X \) of the FFT-processed samples, such that  \(  \ X = (x_0, x_1, \ldots, x_{L-1})\).
\subsubsection{Instruction fine-tuning}
Fine-tuning helps a model to better grasp specific industrial terminologies, fault mechanisms, and operational settings, thereby enhancing its ability to generate accurate and contextually relevant answers in the target task.  Instruction tuning  \citep{Wei2021FinetunedLM, Ouyang2022TrainingLM} is a fully supervised fine-tuning technique used to further train LLMs on specific target domains. This approach not only enhances the controllability of LLMs to follow human instructions helpfully and safely but also enables them to adapt their existing knowledge to the nuanced demands of new tasks. In fact, recent studies \citep{Sanh2021MultitaskPT} have shown that instruction-tuned LLMs exhibit strong zero-shot generalizability on unseen tasks, which is a highly valuable trait in industrial fault diagnosis, where limited data and diverse operational conditions pose significant challenges in building a robust, generalized system.

We employ Low-Rank Adaptation (LoRA) \citep{Hu2021LoRALA}, an efficient approach for fine-tuning large models which reduces the computational burden by introducing trainable low-rank matrices into the model’s layers. Instead of updating the full weight matrix \( W \in \mathbf{R}^{H \times H} \) in each layer, LoRA decomposes the weight update into the sum of a frozen base matrix \( W \) and a low-rank matrix \( \Delta W \). The decomposition is given by:
\begin{equation}
    W' = W + \Delta W
    \label{eq:lora}
\end{equation} 
where \( \Delta W = A B^T \), with \( A \in \mathbf{R}^{H \times R} \) and \( B \in \mathbf{R}^{H \times R} \), and \( R \ll \ H \). During training, only the low-rank matrices \( A \) and \( B \) are updated, while the original weight matrix \( W \) remains frozen, significantly reducing the number of trainable parameters. In the forward pass, the  original intermediate calculation \( \mathbf{h} = W \cdot I \) is modified as:
\begin{equation}
   \mathbf{h} = W \cdot I + A B^\top \cdot I
    \label{eq:forwardpass} 
\end{equation}

\noindent where \( \alpha \) is a scaling factor that controls the contribution of the low-rank update \( \Delta W = A B^T \). This ensures that the low-rank adaptation integrates smoothly with the pre-trained model, preventing the low-rank update from overpowering the original weights. By freezing the main weights and updating only the low-rank matrices, LoRA achieves efficient fine-tuning with minimal computational overhead. 

\subsubsection{Post-processing and evaluation}

Since the LLMs produce predictions in natural language, to evaluate the performance of the LLM using standard DL and ML metrics, we map the textual predictions and corresponding true labels into a numerical or categorical format by defining a mapping function $\phi$ such that: 

\begin{equation}
    \phi : \hat{t}_{text} \rightarrow \hat{t}, \quad \hat{t} \in T
\label{eq；pred}
\end{equation}
\noindent where $\hat{t}_{text}$ is the LLM's predicted text, and  $t_{text}$ is the  true label text.  Similarly, the true label $t$ is mapped as:
\begin{equation}
    \phi : t_ {text}\rightarrow t, \quad t \in T
\label{eq:t-pred}
\end{equation}
We can then evaluate the model's performance by comparing the predicted class $\hat{t}$ against the true class $t$, using metrics such as Accuracy (Acc), Precision (Prec), Recall (Rec), F1-Score (F1), and Confusion Matrix (MC).

\section{Experiments}
\subsection{Dataset source and tasks}
\textbf{Dataset:} In this study, we use the Case Western Reserve University (CWRU) dataset\footnote{http://csegroups.case.edu/bearingdatacenter/home}, which includes vibration signals from ball bearings in both healthy and faulty states. Faults were deliberately introduced into the bearings using electro-discharge machining (EDM), generating single-point defects of varying sizes (0.007, 0.014, and 0.021 inches in diameter) on critical bearing components such as the inner race, outer race, and rolling elements. Then, vibration data were systematically recorded via accelerometers mounted at both the drive end and fan end of the motor housing at sampling rates of 12KHz and 48KHz, and under four operational conditions, including motor loads of 0HP, 1HP, 2HP, and 3HP, and speeds ranging from 1797 to 1730 RPM. 

We use the data collected at a 12KHz sampling rate for both drive end and fan end bearings. 
The data collected from both the drive end and fan end are pre-processed as described in Section \ref{set:3.2.1}. A single label was assigned to each bearing condition, irrespective of fault size, as shown in Table \ref{table:2}. For instance, all inner race faults with diameters ranging from 0.007 inches to 0.021 inches were grouped under the "Inner Race Fault" (IRF) label. The resulting dataset contains four types of faults—Normal (NO), Inner Race Fault (IRF), Outer Race Fault (ORF), and Rolling Element Fault (REF).
Additionally, a more detailed dataset was constructed following traditional settings of DL and ML based fault diagnosis, where each fault size is treated as a distinct fault type. The detailed configuration is shown in Table \ref{table:3}.

\begin{table}[h!]
\begin{center}
\scriptsize
\setlength{\tabcolsep}{10pt}
\renewcommand{\arraystretch}{2}
\caption{Dataset configuration for FFT and Statistically Processed Data. The table outlines the samples corresponding to each working condition, which form subsets for either the drive end or the fan end. These subsets are designated as 0HPDE, 1HPDE, 2HPDE, and 3HPDE for the drive end, where "xHP" indicates the working condition and "DE" denotes the drive end. Similarly, for the fan end, the subsets are labelled 0HPFE, 1HPFE, 2HPFE, and 3HPFE, with "FE" representing the fan end.}\label{table:2}%
\begin{tabular}{@{}ccccccccccccc@{}}
\toprule
Work condition  & \multicolumn{3}{c}{0HP/1797RPM}      & \multicolumn{3}{c}{1HP/1772RPM}      & \multicolumn{3}{c}{2HP/1750RPM}      & \multicolumn{3}{c}{2HP/1750RPM}      \\ \hline
Fault size/inch & 0.007       & 0.014      & 0.021     & 0.007       & 0.014      & 0.021     & 0.007       & 0.014      & 0.021     & 0.007       & 0.014      & 0.021     \\
\midrule
Label           & \multicolumn{3}{c}{NO/IRF/ ORF/ REF} & \multicolumn{3}{c}{NO/IRF/ ORF/ REF} & \multicolumn{3}{c}{NO/IRF/ ORF/ REF} & \multicolumn{3}{c}{NO/IRF/ ORF/ REF} \\
No of samples   & \multicolumn{3}{c}{230/690/690/690}  & \multicolumn{3}{c}{230/690/690/690}  & \multicolumn{3}{c}{230/690/690/690}  & \multicolumn{3}{c}{230/690/690/690}  \\
Total           & \multicolumn{3}{c}{2300}             & \multicolumn{3}{c}{2300}             & \multicolumn{3}{c}{2300}             & \multicolumn{3}{c}{2300}             \\
Subsets         & \multicolumn{3}{c}{0HPDE/0HPFE}      & \multicolumn{3}{c}{1HPDE/1HPFE}      & \multicolumn{3}{c}{2HPDE/2HPFE}      & \multicolumn{3}{c}{3HPDE/3HPFE} \\
\bottomrule
\end{tabular}
\end{center}
\end{table}

\begin{table}[h!]
\begin{center}
\scriptsize
\setlength{\tabcolsep}{18pt}
\renewcommand{\arraystretch}{1.5}
\caption{Dataset detailed configuration}\label{table:3}%
\begin{tabular}{@{}cccc@{}}
\toprule
Labels              & Fault size/inches & Working condition            & No of samples \\
\midrule
NO                  & Non               & \multirow{4}{*}{0HP/1797RPM} & 230           \\
IFR1/ ORF1/ REF1    & 0.0070            &                              & 690           \\
IFR2/ ORF2/ REF2    & 0.0014            &                              & 690           \\
IFR3/ ORF3/ REF3    & 0.021             &                              & 690           \\
NO                  & Non               & \multirow{4}{*}{1HP/1772RPM} & 230           \\
IFR1/ ORF1/ REF1    & 0.0070            &                              & 690           \\
IFR2/ ORF2/ REF2    & 0.0014            &                              & 690           \\
IFR3/ ORF3/ REF3    & 0.021             &                              & 690           \\
NO                  & Non               & \multirow{4}{*}{2HP/1750RPM} & 230           \\
IFR1/ ORF1/ REF1    & 0.0070            &                              & 690           \\
IFR2/ ORF2/ REF2    & 0.0014            &                              & 690           \\
IFR3/ ORF3/ REF3    & 0.021             &                              & 690           \\
NO                  & Non               & \multirow{4}{*}{2HP/1730RPM} & 230           \\
IFR1/ ORF1/ REF1    & 0.0070            &                              & 690           \\
IFR2/ ORF2/ REF2    & 0.0014            &                              & 690           \\
IFR3/ ORF3/ REF3    & 0.021             &                              & 690           \\
Total No of samples & \multicolumn{2}{c}{}                             & 9200         \\
\bottomrule
\end{tabular}
\end{center}
\end{table}

\textbf{Tasks:} We designed a series of experimental tasks under different settings to comprehensively evaluate the performance of LLMs for fault diagnosis in terms of adaptability to various operational conditions (0HP, 1HP, 2HP, etc.) and generalizability, by analyzing data across different components of the machine (Drive end and Fan end).

\textbf{Task 1: Traditional fault diagnosis settings.} 

In this task, we follow the common experimental settings of fault diagnosis. Specifically, we combine the subsets collected from the drive end (0HPDE, 1HPDE, 2HPDE, and 3HPDE) into a dataset named CWRUfft-DE for FFT-processed data and CWRUst-DE for statistically processed data. Similarly, the subsets from the fan end (0HPFE, 1HPFE, 2HPFE, and 3HPFE) are merged into datasets called CWRUfft-FE and CWRUst-FE. For each merged dataset, 10\% is reserved for evaluation.

\textbf{Task 2: Cross-dataset settings.} 
In this task, we assess the adaptation capabilities of LLMs by conducting domain-specific fine-tuning and cross-domain evaluation as follows.  First, we fine-tune LLMs on the data collected under certain operational conditions (source domain). Specifically, we utilize 0HPDE subset collected from the drive end operating at 0HP load and speed of 1797rpm to train LLMs. In this phase, 10\% of 0HPDE data is used for evaluation which is similar to any traditional fault diagnosis settings.

We then investigate the adaptability of these fine-tuned models using subsets from other operational conditions (target domains) but within the same machine component (drive end). These subsets are 1HPDE, 2HPDE, and 3HPDE, which are collected under loads of 1HP, 2HP, and 3HP respectively. 

To further investigate LLMs generalizability across distinct machine components, we also evaluate the fine-tuned models utilizing 0HPFE and 1HPFE subsets from the fan end (target domain). This task settings enables us to examine the model’s generalizability from the drive end training data (source domain) to fan-end test data in cross-machine components evaluation.

\textbf{Task 3: Overall evaluation.} In this task, we integrate all available data from both the drive end and the fan end into a unified dataset named CWRUfft-all and CWRUst-all for FFT processed data and statistically processed data, respectively. 90\% of the data is used for fine-tuning and a proportion of 10\% is allocated for evaluation. By evaluating the model on this extensive dataset, we assess the model's overall effectiveness and robustness in predicting faults across a broad spectrum of operational conditions and different machine components. 
\subsection{Models}
To fully validate the effectiveness of our FD-LLM on the test dataset, we select three categories of models:

\begin{itemize}
    \item {\bf ML Algorithm:} For the traditional ML category, we use Support Vector Machine (SVM) which is a widely used ML algorithm in machine health state identification especially for the essential elements of rotating machinery, such as gears, rolling bearings, and motors. 

\item {\bf DL Algorithm:} For the DL category, we use the one-dimensional convolutional neural network (1D-CNN) model, which is a popular yet efficient DL model used in many fault diagnosis frameworks and has proven its capability of producing accurate predictions. Specifically, we used the WDCNN\citep{Zhang2017AND} model after adding minor changes to the network’s structure, such as rescaling the kernel size of the first convolutional layer from 64 to 16 and reducing the stride to 2, to meet the length of the samples in our data. 

\item {\bf Open-source LLM:} For the LLM category, we use four leading open-source LLMs namely, Llama3-8B, Llama3-8B instruct, Qwen1.5-7B, and Mistral-7B. The selection of these models is based on their wide range of applications and their capacities in handling numerical data.
\end{itemize}

\subsection{Settings}
 Since both the ML and DL algorithms can only take in numerical data, we used only the statistical data without incorporating any text for the ML algorithm, and only the FFT dataset for the DL algorithm. Although both ML and DL models have the ability to process either FFT or statistical data, our selection is based on best practices and the strengths of each approach: DL models are most suitable for analyzing high-dimensional data like FFT vectors that capture the frequency-domain characteristics of vibration signals, while ML models frequently benefit from compact, high-level statistical features. In contrast, LLMs are able to incorporate both the textual data and the vibration data either in the form of statistical quantities or encoded FFT vectors.

Specifically, the hyperparameters used in this study are as follows: batch size 2, LoRA rank 4, cosine lr scheduler, learning rate 1e-4, bf16, and NVIDIA A10 for all training processes. We trained 3 epochs for all LLMs experiments.
\subsection{Metrics}
We employ multiple standard DL and ML evaluation metrics, including Accuracy, Precision, Recall, and F1-Score, to evaluate the LLM's classification performance. 
For conciseness, we will focus on presenting the accuracy and F1-score in the following discussion, with results for the other metrics provided in Appendix \ref{sec:B}.

\subsection{Results and discussions}

\subsubsection{Task 1: Traditional fault diagnosis settings}
 Table \ref{table:4} shows the results of the traditional fault diagnosis settings. All models were evaluated using the statistically processed data (CWRUst-DE) and FFT-processed data (CWRUfft-DE) from the drive end, as well as CWRUst-FE and CWRUfft-FE from the fan end.

\begin{table}[h!]
\centering
\scriptsize
\renewcommand{\arraystretch}{1.8}
\caption{Evaluation results of all models under combined datasets from drive end and fan end subsets}\label{table:4}
\resizebox{\linewidth}{!}{%
\begin{tabular}{ccccccccc}
\toprule
\multicolumn{1}{c}{\multirow{3}{*}{Model}} & \multicolumn{4}{c}{Drive end}                           & \multicolumn{4}{c}{Fan end}                           \\ 
\cline{2-9} 
\multicolumn{1}{c}{}  
   & \multicolumn{2}{c}{CWRUst-DE}         
   & \multicolumn{2}{c}{CWRUfft-DE}                    & \multicolumn{2}{c}{CWRUst-FE}                     & \multicolumn{2}{c}{CWRUfft-FE} \\ 
\cline{2-9} 
\multicolumn{1}{c}{}                       & \multicolumn{1}{c}{Accuracy} & \multicolumn{1}{c}{F1-Score} & \multicolumn{1}{c}{Accuracy} & \multicolumn{1}{c}{F1-Score} & \multicolumn{1}{c}{Accuracy} & \multicolumn{1}{c}{F1-Score} & \multicolumn{1}{c}{Accuracy} & F1-Score        \\ 
\midrule
Llama3    &  {\ul{\textbf{0.9480}}} & {\ul{\textbf{0.9402}}}  & {\ul{\textbf{0.997}}}    & {\ul{\textbf{0.9969}}}  & {\ul{\textbf{0.8467}}}   & {\ul{\textbf{0.8464}}}   & {\ul{\textbf{0.9875}}}  & {\ul{\textbf{0.9874}}} \\
Qwen1.5-7B    & 0.3826 & 0.3711  & 0.762  & 0.7562 & 0.3508   & 0.3127                                 & 0.6725  & 0.6484  \\
Mistral-7B    & 0.3453  & 0.2641   & 0.5390 & 0.4951  & 0.3347  & 0.2530                              & 0.3504  & 0.3059 \\
Llama3-instruct    & {\ul{\textbf{0.9521}}} & {\ul{\textbf{0.9520}}} & {\ul{\textbf{0.998}}} & {\ul{\textbf{0.998}}}  & {\ul{\textbf{0.9097}}} & {\ul{\textbf{0.9094}}} & {\ul{\textbf{0.9975}}} & 
                    {\ul{\textbf{0.9974}}} \\
SVM                & {\ul{\textbf{0.9739}}} & {\ul{\textbf{0.9739}}}& N/A  & N/A & 0.8733 & 0.8760                  & N/A  & N/A  \\
WDCNN            & N/A & N/A & {\ul{\textbf{0.9928}}} & {\ul{\textbf{0.9911}}} & N/A        & N/A & {\ul{\textbf{0.9916}}} & {\ul{\textbf{0.9910}}} \\ 
\bottomrule
\end{tabular}}
\end{table}

\textbf{Statistically processed data.} The evaluation results on the statistical datasets CWRUst-DE and CWRUst-FE reveal significant variations in the performance across different models. Llama3 and Llama3-instruct exhibited relatively satisfactory results on the drive end data (CWRUst-DE), in which Llama3 obtained accuracy and F1-score of 0.9480 and 9420, respectively, and Llama3-instruct demonstrated relatively higher metrics, with an accuracy of 0.9521 and an F1-score of 0.9520. However, on the fan end data (CWRUst-FE), both Llama3 and Llama3-instruct showed a substantial decline in performance, with decreases of 10\% and 5\% in accuracy, respectively. Moreover, the other LLMs, Qwen1.5-7B and Mistral7B\_v0.2 displayed consistently low accuracy and F1-scores, indicating their failure to generalize effectively across both datasets. Additionally, the ML method represented by SVM achieved the best performance on CWRUst-DE but yielded relatively low accuracy and F1-scores on CWRUst-FE.

\textbf{FFT processed data.} Llama3 and Llama3-instruct achieved perfect fault diagnostic accuracy on FFT-processed data from both the drive end (CWRUfft-DE) and the fan end (CWRUfft-FE). Conversely, Qwen1.5-7B and Mistral7B\_v0.2 continued to exhibit low accuracy and F1-scores, demonstrating poor generalization across both datasets. However, DL models, represented by WDCNN, achieved competitive results. These evaluation results demonstrate that FFT-processed data provides richer information from which the models can effectively learn, leading to superior performance in fault diagnosis compared to statistically processed data.

In answering the question “Are LLMs valid fault diagnosis tools?”, the results have indeed shown that LLMs such as LLama3 and LLama3-Instruct exhibited robust fault diagnosis performance. This effectiveness can be attributed to their ability to interpret numerical data when appropriately pre-processed, leverage domain knowledge, and recognize complex patterns across diverse inputs. In fact, Llama models take numerical tokenization into account during their pre-training stage, enabling effective handling of complex token patterns-- an advantage that likely contributes to its higher fault diagnosis accuracy as compared to models like Mistral and Qwen1.5\citep{Touvron2023Llama2O}. In contrast, Mistral prioritizes modularity and computational efficiency over detailed numerical pattern learning, while Qwen1.5 focuses on extended context length and multilingual robustness rather than specific adaptations for numerical data processing. Additionally, the models achieved near 100\% accuracy and F1-score on FFT-processed data, highlighting that the FFT transformations provide richer, more informative representations and thus enable LLMs to extract meaningful insights, resulting in superior fault diagnosis performance compared to statistical features alone.

\subsubsection{Task 2: Cross-dataset evaluation}
In this task, we conducted a zero-shot evaluation for all models to assess the generalization abilities of LLMs compared to ML and DL-based fault diagnosis models. Specifically, we trained all models on the 0HPDE subset and carried out the evaluation as follows:

\renewcommand{\labelenumi}{(\arabic{enumi})}
\begin{enumerate}
    \item {\it Within the same subset:}  To evaluate within the same subset, we use 10\% of 0HPDE for evaluation following common fault diagnosis experimental settings;
    \item  {\it Across operational conditions:} To evaluate within the same machine component (drive end) but across operational conditions (target domains),  we assess the models under different operational conditions using subsets from the drive end (1HPDE, 2HPDE, and 3HPDE);
    \item {\it Across machine components:} To evaluate across different machine components (target domains), we evaluate all models using two subsets from the fan end (0HPFE and 1HPFE).
\end{enumerate}
Tables \ref{table:5} and \ref{table:6} display the evaluation results of all models using both statistical and FFT-processed data, respectively.

\textbf{Statistically processed data:} As presented in Table \ref{table:5}, the evaluation results revealed low levels of generalization and adaptability across different target domains for the tested models. This is likely due to the nature of statistical representations, which capture only global properties of vibration signals and may overlook subtle changes or local patterns essential for adapting to different operational conditions. Consequently, the evaluation indicates that statistical representations do not improve the generalization capabilities of LLMs.

\renewcommand{\labelenumi}{(\arabic{enumi})}
\begin{enumerate}
    \item {\it Within the same subset:} Llama-3 demonstrated the strongest performance compared to other LLMs, achieving the highest accuracy (97.62\%) and F1-score (96.93\%) on 0HPDE data. Similarly, SVM showed a strong diagnostic performance.
    \item {\it Across operational conditions:} The diagnostic performance of all models dramatically declined as they were exposed to data from increasingly divergent operational conditions. The best accuracy achieved on 1HPDE by SVM is 77.60\% and further dropped to 71.30\% on 2HPDE. 

\item{\it Across-machine components:} None of the models, including SVM, were able to generalize well to data from different machine components, indicating poor adaptability to unseen conditions from the fan end.
\end{enumerate}

\begin{table}[h!]
\begin{center}
\scriptsize
\renewcommand{\arraystretch}{1.8}
\caption{ The cross-dataset evaluation results of all models using statistically processed data}
\label{table:5}
\resizebox{1\textwidth}{!}{
\begin{tabular}{@{}ccccccccccc@{}}
\toprule
\multirow{2}{*}{Data} &    \multicolumn{2}{c}{Llama3-8B}    & \multicolumn{2}{c}{SVM}   & \multicolumn{2}{c}{Qwen1.5-7B} & \multicolumn{2}{c}{Llama3-instruct}    & \multicolumn{2}{c}{Mistral-7B} \\ \cline{2-11} 
                           & Accuracy & F1-Score             & Accuracy & F1-Score       & Accuracy  & F1-Score              & Accuracy  & F1-Score                   & Accuracy   & F1-Score      \\
\midrule
0HPDE     & {\ul{\textbf{0.9762}}} & {\ul{\textbf{0.9693}}} & {\ul{\textbf{0.9760}}} & {\ul{\textbf{0.9759}}} & 0.8102 & 0.8096  & {\ul{\textbf{0.8956}}} & {\ul{\textbf{0.8874}}} & 0.665 & 0.6415 \\
1HPDE                 & 0.7230  & 0.7170                & 0.7760 & 0.7734                & 0.6647  & 0.6497               & 0.7091    & 0.7016                     & 0.3785         & 0.3698        \\
2HPDE                 & 0.6669  & 0.6273                & 0.7130    & 0.6972                & 0.6665& 0.6284             & 0.6630      & 0.6219                     & 0.4034     & 0.3904        \\
3HPDE                 & 0.6678  & 0.6166                & 0.7826    & 0.7646                & 0.6552  & 0.5993            & 0.6808    & 0.6301                      & 0.3614         & 0.3408    \\
0HPFE                 & 0.6552    & 0.6263                & 0.5043  & 0.4998                & 0.6143   & 0.5906           & 0.6095    & 0.5769                       & 0.3634         & 0.3515     \\
1HPFE                 & 0.3630   & 0.3329                & 0.4956   & 0.5020                & 0.3508     & 0.3127         & 0.3521     & 0.3222                       & 0.3697         & 0.3538    \\
\bottomrule
\end{tabular}}
\end{center}
\end{table}

\begin{table}[h!]
\begin{center}
\scriptsize
\renewcommand{\arraystretch}{1.8}
\caption{ The cross-dataset evaluation results of all models using FFT-processed data}
\label{table:6}
\resizebox{1\textwidth}{!}{
\begin{tabular}{ccccccccccc}
\toprule
\multicolumn{1}{c}{\multirow{2}{*}{Data}} & \multicolumn{2}{c}{Llama3-8B}                                & \multicolumn{2}{c}{WDCNN}  & \multicolumn{2}{c}{Llama3-instruct}                     & \multicolumn{2}{c}{Qwen1.5-7B} & \multicolumn{2}{c}{Mistral-7B}\\
\cline{2-11} 
\multicolumn{1}{c}{} & \multicolumn{1}{c}{Accuracy} & \multicolumn{1}{c}{F1-Score} & \multicolumn{1}{c}{Accuracy} & \multicolumn{1}{c}{F1-Score} & \multicolumn{1}{c}{Accuracy} & \multicolumn{1}{c}{F1-Score} & \multicolumn{1}{c}{Accuracy} & \multicolumn{1}{c}{F1-Score} & \multicolumn{1}{c}{Accuracy} & F1-Score \\
\midrule

0HPDE      
        & {\ul{\textbf{1.0}}} & {\ul{\textbf{1.0}}} & {\ul{\textbf{0.999}}} & {\ul{\textbf{0.9989}}} & {\ul{\textbf{1.0}}} & {\ul{\textbf{1.0}}} & 
        {\ul{\textbf{0.968}}} & {\ul{\textbf{0.9680}}} & 0.548  & 0.5518   \\
        
1HPDE  
        & {\ul{\textbf{0.986}}}& {\ul{\textbf{0.9859}}} & 0.927 & 0.9264 & 
        {\ul{\textbf{0.998}}}& {\ul{\textbf{0.9979}}} & 0.9304 & 0.9297                  & 0.5088 & 0.5059   \\
2HPDE 
        & {\ul{\textbf{0.9376}}}& {\ul{\textbf{0.9359}}} & 0.912& 0.9109                        & {\ul{\textbf{0.9648}}} & {\ul{\textbf{0.9643}}} & 0.8964     & 0.8961 & 0.4936   & 0.4894   \\
        
3HPDE                                      
            & {\ul{\textbf{0.9376}}}& {\ul{\textbf{0.9359}}}& 0.848 & 0.8455              & {\ul{\textbf{0.9648}}}  & {\ul{\textbf{0.9643}}} & 0.8964 & 0.8961    & 0.4936 & 0.4894   \\
0HPFE                                    
            & 0.4377  & 0.4100  & 0.479 & 0.4539 & 0.529 & 0.4757                        & 0.6565 & 0.6322   & 0.4663  & 0.4722   \\
            
1HPFE                                     
            & 0.433 & 0.3876  & 0.396  & 0.3539  & 0.4755  & 0.4099                        & 0.586 & 0.5186  & 0.4805  & 0.4840   \\ 
\bottomrule
\end{tabular}}
\end{center}
\end{table}

\textbf{FFT-processed data:} Table \ref{table:6} summarises all models' evaluation results on FFT data and shows that Llama3 and Llama3-instruct exhibited the most satisfactory results. 
\renewcommand{\labelenumi}{(\arabic{enumi})}
\begin{enumerate}
    \item {\it Within the same subset:} Amongst all the evaluated LLMs, Llama3 and Llama3-instruct delivered the best results, achieving perfect accuracy and F1-score (100\%). Qwen1.5 also demonstrated relatively strong performance, while Mistral showed the lowest accuracy and F1-scores at 54.8\% and 55.18\%, respectively. On the other hand, WDCNN exhibited strong diagnostic accuracy at 99.9\%, outperforming both Qwen1.5 and Mistral.
    
 \item {\it Across operational conditions:} Llama3 and Llama3-instruct maintained robust performance, with Llama3-instruct showing significant superiority on 2HPDE and 3HPDE, demonstrating strong adaptation and generalization to unseen conditions. In contrast, Qwen1.5's performance declined rapidly as operational conditions became more divergent, while Mistral continued to underperform. WDCNN displayed acceptable yet competitive results, though its adaptation was expectedly limited. DL models like WDCNN often struggle with new operational conditions or equipment due to the distributional discrepancies between training and test data, which explains the performance degradation, particularly on 2HPDE and 3HPDE. 
 \item {\it Across-machine components:} The results indicate a significant performance decline for all models when applied to data from different machine components, underscoring poor generalization to new mechanical devices. 
    
\end{enumerate}
 
\subsubsection{Task 3: Overall evaluation}
For overall evaluation, we constructed one comprehensive dataset that encompassed all the subsets from the drive end and fan end, using 90\% of this dataset for training all models. We then evaluate the models using the remaining 10\% of the data. The evaluation was conducted using both statistically processed data (CWRUst-all) and FFT processed data (CWRUfft-all). 

Based on the results presented in Table \ref{table:7},   Llama-3-8B and FF-DM achieved the highest performance, demonstrating strong generalization across both the statistically processed data (CWRUst-all) and FFT processed data (CWRUfft-all). In contrast, models such as Qwen1.5-7B and Mistral7B\_v0.2 performed poorly. probably due to their design choices that position them as strong general-purpose models but less specialized for tasks requiring complex numerical data interpretation. Traditional machine learning methods, such as SVM, also did not perform  well. 

\begin{table}[h!]
\begin{center}
\scriptsize
\setlength{\tabcolsep}{20pt}
\renewcommand{\arraystretch}{1.8}
\caption{ Evaluation results of all models using all data from the drive and fan end. In the table, CWRUst-all represents all statistically processed data from both the drive and fan end, while CWRUfft-all denotes the FFT-processed data from both the drive and fan end  }
\label{table:7}
\resizebox{1\textwidth}{!}{
\begin{tabular}{ccccc}
\toprule
\multirow{2}{*}{model} & \multicolumn{2}{c}{CWRUst-all}                & \multicolumn{2}{c}{CWRUfft-all}               \\ \cline{2-5} 
                       & Accuracy              & F1-Score              & Accuracy              & F1-Score              \\
\midrule
Llama-3-8B             & {\ul {\textbf{0.9480}}} & {\ul {\textbf{0.9407}}} & {\ul {\textbf{0.99}}}   & {\ul {\textbf{0.990}}}  \\
Qwen1.5-7B             & 0.5309                & 0.5357                & 0.3766                & 0.330                 \\
Mistral-7B        & 0.2843                & 0.2010                & 0.3857                & 0.3233                \\
Llama3-instruct        & {\ul {\textbf{0.9538}}} & {\ul {\textbf{0.9537}}} & {\ul {\textbf{0.9988}}} & {\ul {\textbf{0.9988}}} \\
SVM                    & 0.9445                & 0.9446                & N/a                   & N/a                   \\
WDCNN                  & N/a                   & N/a                   & {\ul {\textbf{0.9641}}} & {\ul {\textbf{0.9627}}} \\
\bottomrule

\end{tabular}}
\end{center}
\end{table}

However, DL models, represented by WDCNN, exhibited strong performance on the FFT processed data, achieving an accuracy of 96.41\% and an F1-score of 96.27\%. This result indicates that DL models can still be highly effective. Nevertheless, LLMs possess an advantage in their ability to integrate not only numerical data but also textual information that describes machine operational conditions and specifications. This capability allows LLMs to contextualize vibration signals with additional insights, enabling them to capture more nuanced patterns and relationships in the data, thereby achieving relatively higher performance than ML and DL models, especially when deployed for fault diagnosis on different devices.

\subsection{Ablation study}
We also performed an ablation study to systematically evaluate the influence of different dataset configurations on the performance of LLMs for fault diagnosis. The objective is to understand how various settings and data preprocessing techniques affect the overall effectiveness of LLMs. 

First, we investigate the impact of incorporating machine specifications into the input prompts, focusing on the performance of Llama3 and Llama3-instruct. This evaluation is conducted using the comprehensive datasets outlined in Task 3. Specifically, we compare the performance of each model with and without machine specifications included in the input prompts.
Then, we examine the effect of dataset labelling configurations. We compare the models' performance using the dataset structure presented in Table \ref{table:3} with an alternative configuration where only one label per fault is used, regardless of fault size. This helps us understand whether detailed labelling or simplified labelling is more beneficial for fault diagnosis tasks. The evaluation results are as follows:

\begin{itemize}
    \item \textbf{Impact of Incorporating Machine Specifications:} As shown in Table \ref{table:8}, including machine specifications in the input prompts significantly improves the performance of both Llama3 and Llama3-instruct on statistically processed data, yielding a 20\% and 11\% increase in accuracy, respectively. However, when machine specifications are incorporated into FFT-processed data, the performance gains are less noticeable, as both models already exhibit high accuracy and F1-scores on this data.

    \item \textbf{Effect of dataset labelling configurations:} From Table \ref{table:9}, we found out that detailed labelling of the fault type according to their malfunction size did not significantly impact the performance of LLMs. The models had remained robust, demonstrating that LLMs, besides identifying the faults, can determine the fault severity effectively.
\end{itemize}

\begin{table}[h]
\begin{center}
\scriptsize
\renewcommand{\arraystretch}{1.8}
\setlength{\tabcolsep}{18pt}
\caption{ Impact of incorporating machine specifications in the instruction prompt using statistical and FFT-processed data}

\label{table:8}
\resizebox{1\textwidth}{!}{
\begin{tabular}{ccccc}
\toprule
\multirow{2}{*}{Data}                         & \multicolumn{2}{c}{Llama3-8B}                 & \multicolumn{2}{c}{Llama3-instruct}           \\ \cline{2-5} 
                                              & Accuracy              & F1-Score              & Accuracy              & F1-Score              \\
                                              
\midrule
Statistical   data (No machine specification) & 0.7467                & 0.7455                & 0.8451                & 0.8449                \\
Statistical   data (machine   specification)  & {\ul {\textbf{0.9480}}} & {\ul {\textbf{0.9407}}} & {\ul {\textbf{0.9555}}} & {\ul {\textbf{0.9556}}} \\
FFT   data (No machine specification)         & {\ul {\textbf{0.975}}}  & {\ul {\textbf{0.9750}}} & {\ul {\textbf{0.99}}}   & {\ul {\textbf{0.990}}}  \\
FFT   data (machine specification)            & {\ul{\textbf{0.99}}}   & {\ul {\textbf{0.990}}}  & {\ul {\textbf{0.9988}}} & {\ul {\textbf{0.9988}}}\\
\bottomrule
\end{tabular}}
\end{center}
\end{table}
\vspace{-20pt}

\begin{table}[h]
\begin{center}
\scriptsize
\renewcommand{\arraystretch}{1.8}
\setlength{\tabcolsep}{18pt}
\caption{ Effect of dataset labelling configurations using all FFT-processed data from drive end and fan end}
\label{table:9}
\resizebox{1\textwidth}{!}{
\begin{tabular}{ccccc}
\toprule
\multirow{2}{*}{Data} & \multicolumn{2}{c}{Llama3-8B}                 & \multicolumn{2}{c}{Llama3-instruct}           \\ \cline{2-5} 
                      & Accuracy              & F1-Score              & Accuracy              & F1-Score              \\
\midrule
CWRUfft-all-10labels  & {\ul {\textbf{0.9910}}} & {\ul {\textbf{0.9904}}} & {\ul {\textbf{0.9944}}} & {\ul {\textbf{0.9944}}} \\
CWRUfft-all-4labels   & {\ul {\textbf{0.99}}}   & {\ul {\textbf{0.990}}}  & {\ul {\textbf{0.9988}}} & {\ul {\textbf{0.9988}}}\\
\bottomrule
\end{tabular}}
\end{center}
\end{table}
\FloatBarrier

\section{Conclusion}
This study presents FD-LLM, a novel framework that bridges the gap between fault diagnosis and advanced language modeling through three key steps: data pre-processing, instruction fine-tuning, and post-processing. In the pre-processing phase, the challenge of aligning vibration signal modalities with LLM input formats was addressed by encoding the vibration signals into text. Two encoding methods were employed: string-based tokenization of FFT-processed signals and statistical summaries derived from both time and frequency domains. The second step involves instruction fine-tuning using LoRA, which allows for efficient adaptation of LLMs to fault diagnosis tasks. In the final post-processing step, the LLM-generated predictions were mapped to numerical labels for the calculation of evaluation metrics in the assessment of model performance.

Our extensive experiments have validated the effectiveness of FD-LLM in various fault diagnosis scenarios. Models such as Llama3 and Llama3-instruct demonstrated exceptional diagnostic performance in all settings, particularly when utilizing FFT-processed data. These models also exhibited strong adaptability, achieving high accuracy in diagnosing faults under new operational conditions. However, performance was lower when the models were tasked with diagnosing faults across different machine components, revealing a challenge in cross-component generalization. 
 
In summary, FD-LLM has showcased the considerable potential of utilizing LLMs for intelligent fault diagnosis across a range of diagnostic scenarios. On the other hand, our experiments have highlighted that future research should focus on enhancing cross-component adaptability to improve the system's robustness and reliability. One promising direction for achieving this would be the incorporation of reasoning intelligence into the fault diagnosis process such as chain-of-thought (CoT) \citep{kim2023cot} or the more fine-grained Process-Supervised Reward Model (PRM)\citep{ma2023let}, which would guide the LLMs through a structured diagnostic process to systematically analyze vibration signals, calculate characteristic fault frequencies step by step, and progressively generate more accurate fault predictions.  

\bibliographystyle{unsrtnat}
\bibliography{reference}  

\appendix
\section{Appendix A. Statistical Features Calculation}
\label{sec:app-A}
Table \ref{tab:summary_features} presents the features extracted from time and frequency domains and their calculation formulas, where \(x_{j,k}(n)\) represents the $k^{th}$ segment from the $j^{th}$ signal.  \(|X_{j,k}(m)|\) denotes the magnitude of the FFT output, \(\mu_{|X|}\)  is the mean (average) value of the magnitudes, and \(\sigma_{|X|}\) represents the standard deviation of the magnitudes.

\begin{table}[h]
\centering
\setlength{\tabcolsep}{18pt}
\renewcommand{\arraystretch}{2.5}
\caption{Summary of features and formulas used in time and frequency domains.}\label{tab:formulas}
\resizebox{0.9\textwidth}{!}{
\begin{tabular}{lll}
\toprule
\textbf{Domain} & \textbf{Feature} & \textbf{Formula for Segment $k$ from Signal $j$} \\ 
\midrule
\multirow{10}{*}{\textbf{Time}} 
& Mean                              & \( \mu_{j,k} = \frac{1}{L} \sum_{n=0}^{L-1} x_{j,k}(n) \)                                                                   \\
& RMS                               & \( \text{RMS}_{j,k} = \sqrt{\frac{1}{L} \sum_{n=0}^{L-1} x_{j,k}(n)^2} \)                                                   \\
& Standard Deviation                & \( \sigma_{j,k} = \sqrt{\frac{1}{L} \sum_{n=0}^{L-1} \left( x_{j,k}(n) - \mu_{j,k} \right)^2} \)                            \\
& Crest Factor                      & \( \text{CF}_{j,k} = \frac{\max |x_{j,k}(n)|}{\text{RMS}_{j,k}} \)                                                          \\
& Skewness                          & \( \text{Skew}_{j,k} = \frac{\frac{1}{L} \sum_{n=0}^{L-1} \left( x_{j,k}(n) - \mu_{j,k} \right)^3}{\sigma_{j,k}^3} \)       \\
& Shape Factor                      & \( \text{SF}_{j,k} = \frac{\text{RMS}_{j,k}}{\frac{1}{L} \sum_{n=0}^{L-1} |x_{j,k}(n)|} \)                                  \\
& Kurtosis                          & \( \text{Kurt}_{j,k} = \frac{\frac{1}{L} \sum_{n=0}^{L-1} \left( x_{j,k}(n) - \mu_{j,k} \right)^4}{\sigma_{j,k}^4} \)       \\
& Peak-to-Peak Value                & \( \text{P2P}_{j,k} = \max x_{j,k}(n) - \min x_{j,k}(n) \)                                                                  \\
& Energy Factor                     & \( \text{EF}_{j,k} = \frac{\sum_{n=0}^{L-1} x_{j,k}(n)^2}{\left( \sum_{n=0}^{L-1} |x_{j,k}(n)| \right)^2} \)                \\
& Impulse Factor                    & \( \text{IF}_{j,k} = \frac{\max |x_{j,k}(n)|}{\frac{1}{L} \sum_{n=0}^{L-1} |x_{j,k}(n)|} \) \\

\midrule 
\multirow{5}{*}{\textbf{Frequency}} 
                   & Peak Frequency                    & \( \text{PeakFreq}_{j,k} = \arg\max_m |X_{j,k}(m)| \)                                                                       \\
& Peak-to-Peak Frequency            & \( \text{P2PFreq}_{j,k} = \max |X_{j,k}(m)| - \min |X_{j,k}(m)| \)                                                          \\
& Spectral Kurtosis                 & \( \text{SpecKurt}_{j,k} = \frac{\frac{1}{L} \sum_{m=0}^{L-1} \left( |X_{j,k}(m)| - \mu_{|X|} \right)^4}{\sigma_{|X|}^4} \) \\
& Spectral Bandwidth                & \( \text{SpecBW}_{j,k} = \sqrt{\frac{\sum_{m=0}^{L-1} (m - \mu_f)^2 \cdot |X_{j,k}(m)|}{\sum_{m=0}^{L-1} |X_{j,k}(m)|}} \)  \\ 
& Spectral Skewness                 & \( \text{SpecSkew}_{j,k} = \frac{\frac{1}{L} \sum_{m=0}^{L-1} \left( |X_{j,k}(m)| - \mu_{|X|} \right)^3}{\sigma_{|X|}^3} \) \\
\bottomrule
\end{tabular}}

\label{tab:summary_features}
\end{table}

\section{Appendix B. Additional Evaluation Metrics}
\label{sec:B}
In Tables \ref{table:11}, \ref{table:12}, \ref{table:13}, \ref{table:14}, \ref{table:15}, and \ref{table:16}
, we present additional evaluation metrics (precision and recall) for all experiments from Task 1 to Task 3, along with the ablation study.  
\vspace{-12pt}
\begin{table}[h!]
\begin{center}
\scriptsize
\renewcommand{\arraystretch}{1.8}
\caption{The Evaluation results of all models under combined datasets from drive end and fan end subsets}\label{table:11} 
\resizebox{1\textwidth}{!}{
\begin{tabular}{ccccccccc}
\toprule
\multirow{3}{*}{Model} & \multicolumn{4}{l}{Drive end}                                  & \multicolumn{4}{l}{Fan end}                                    \\ \cline{2-9} 
                       & \multicolumn{2}{l}{CWRUst-DE} & \multicolumn{2}{l}{CWRUfft-DE} & \multicolumn{2}{l}{CWRUst-FE} & \multicolumn{2}{l}{CWRUfft-FE} \\ \cline{2-9} 
                       & Precision       & Recall      & Precision       & Recall       & Precision       & Recall      & Precision       & Recall       \\ \hline
Llama3                 & 0.9405          & 0.9402      & 0.9970          & 0.997        & 0.850           & 0.8467      & 0.9875          & 0.9875       \\
Qwen1.5-7B             & 0.4077          & 0.3826      & 0.7706          & 0.762        & 0.5338          & 0.5369      & 0.6593          & 0.6725       \\
Mistral-7B        & 0.2541          & 0.3453      & 0.5197          & 0.5390       & 0.2037          & 0.3347      & 0.3140          & 0.3504       \\
Llama3-instruct        & 0.9525          & 0.9521      & 0.998           & 0.998        & 0.9104          & 0.9097      & 0.9975          & 0.9975       \\
SVM                    & 0.9744          & 0.9739      & N/A             & N/A          & 0.8997          & 0.8733      & N/A             & N/A          \\
WDCNN                  & N/A             & N/A         & 0.9980          & 0.9928       & N/A             & N/A         & 0.995           & 0.9916  \\      
\bottomrule
\end{tabular}}
\end{center}
\end{table}

\begin{table}[h!]
\begin{center}
\scriptsize
\renewcommand{\arraystretch}{1.8}
\caption{ The cross-dataset evaluation results of all models using FFT processed data}
\label{table:12}
\resizebox{1\textwidth}{!}{
\begin{tabular}{@{}ccccccccccc@{}}
\toprule
\multirow{2}{*}{Data} &    \multicolumn{2}{c}{Llama3-8B}    & \multicolumn{2}{c}{SVM}   & \multicolumn{2}{c}{Qwen1.5-7B} & \multicolumn{2}{c}{Llama3-instruct}    & \multicolumn{2}{c}{Mistral-7B} \\ \cline{2-11} 
                           & Precision       & Recall      & Precision       & Recall      & Precision       & Recall      & Precision       & Recall      & Precision       & Recall      \\
\midrule
0HPDE                 & 0.9651 & 0.9536& 0.9763& 0.9760& 0.9053 & 0.8874& 0.8256 & 0.8074& 0.675 & 0.6615
\\
1HPDE                 & 0.7884& 0.7230& 0.8008& 0.7760& 0.7091& 0.7588& 0.7387& 0.6647& 0.3971& 0.3785
\\
2HPDE                 & 0.770& 0.666& 0.7379& 0.7130& 0.758& 0.6630& 0.7635& 0.6665& 0.4084& 0.4034
\\
3HPDE                 & 0.7877& 0.6678& 0.7856& 0.7826& 0.7783& 0.6808& 0.7538& 0.6552& 0.3728& 0.3614
\\
0HPFE                 & 0.7037& 0.6552& 0.5048& 0.5043& 0.6382& 0.6095& 0.6143& 0.6540& 0.3701& 0.3634
\\
1HPFE                 & 0.3787& 0.3630& 0.5249& 0.49565& 0.3701& 0.3521& 0.4154& 0.350& 0.3625& 0.3697
\\
\bottomrule
\end{tabular}}
\end{center}
\end{table}
\vspace*{-20pt} 
\begin{table}[htbp!]
\begin{center}
\scriptsize
\renewcommand{\arraystretch}{1.8}
\caption{ The cross-dataset evaluation results of all models using statistically processed data}
\label{table:13}
\resizebox{1\textwidth}{!}{
\begin{tabular}{@{}ccccccccccc@{}}
\toprule
\multirow{2}{*}{Data} &    \multicolumn{2}{c}{Llama3-8B}    & \multicolumn{2}{c}{SVM}   & \multicolumn{2}{c}{Qwen1.5-7B} & \multicolumn{2}{c}{Llama3-instruct}    & \multicolumn{2}{c}{Mistral-7B} \\ \cline{2-11} 
                           & Precision       & Recall      & Precision       & Recall      & Precision       & Recall      & Precision       & Recall      & Precision       & Recall      \\
\midrule
0HPDE                 & 1.0& 1.0& 0.999& 0.999& 1.0& 1.0& 0.9683& 0.968& 0.560& 0.548
\\
1HPDE                 & 0.9861& 0.986& 0.9361& 0.927& 0.9980& 0.9979& 0.9308& 0.9304& 0.5114& 0.5088
\\
2HPDE                 & 0.9441& 0.9359& 0.924& 0.912& 0.9666& 0.9648& 0.8966& 0.8964& 0.4929& 0.4936
\\
3HPDE                 & 0.9441& 0.9359& 0.8793& 0.848& 0.9666& 0.9648& 0.8966& 0.8964& 0.4929& 0.4936
\\
0HPFE                 & 0.3933& 0.4100& 0.480& 0.479& 0.529& 0.5634& 0.6734& 0.6565& 0.4914& 0.4663
\\
1HPFE                 & 0.3798& 0.433& 0.3537& 0.398& 0.4351& 0.4755& 0.5550& 0.586& 0.50& 0.4805
\\
\bottomrule
\end{tabular}}
\end{center}
\end{table}

\begin{table}[h!]
\begin{center}
\scriptsize
\setlength{\tabcolsep}{20pt}
\renewcommand{\arraystretch}{1.8}
\caption{ The evaluation results of all models using all data from the drive and fan end. In the table, CWRUst-all represents all statistically processed data from both the drive and fan end, while CWRUfft-all denotes the FFT-processed data from both the drive and fan end  }
\label{table:14}
\resizebox{1\textwidth}{!}{
\begin{tabular}{ccccc}
\toprule
\multirow{2}{*}{model} & \multicolumn{2}{c}{CWRUst-all}                & \multicolumn{2}{c}{CWRUfft-all}               \\ \cline{2-5} 
                       & Precision       & Recall      & Precision       & Recall      \\
\midrule
Llama-3-8B             & 0.9480& 0.9407& 0.99& 0.990
\\
Qwen1.5-7B             & 0.5617& 0.5309& 0.4077& 0.3766
\\
Mistral-7B        & 0.1714& 0.2843& 0.2877& 0.3857
\\
Llama3-instruct        & 0.9538 & 0.9537& 0.9988& 0.9988
\\
SVM                    & 0.9448& 0.9445& N/a& N/a
\\
WDCNN                  & N/a& N/a& 0.955& 0.9627
\\
\bottomrule

\end{tabular}}
\end{center}
\end{table}

\begin{table}[h]
\begin{center}
\scriptsize
\renewcommand{\arraystretch}{1.8}
\setlength{\tabcolsep}{18pt}
\caption{ The Impact of incorporating machine specifications in the instruction prompt using statistical and FFT-processed data}

\label{table:15}
\resizebox{1\textwidth}{!}{
\begin{tabular}{ccccc}
\toprule
\multirow{2}{*}{Data}                         & \multicolumn{2}{c}{Llama3-8B}                 & \multicolumn{2}{c}{Llama3-instruct}           \\ \cline{2-5} 
                                              & Precision       & Recall      & Precision       & Recall      \\
                                              
\midrule
Statistical   data (No machine specification) & 0.7484& 0.7467& 0.8452& 0.8451\\
Statistical   data (machine   specification)  & 0.9483& 0.9480& 0.9555& 0.9556\\
FFT   data (No machine specification)         & 0.9752& 0.975& 0.990& 0.99\\
FFT   data (machine specification)            & 0.99& 0.99& 0.9988& 0.9988\\
\bottomrule
\end{tabular}}
\end{center}
\end{table}
\vspace{-20pt}

\begin{table}[h]
\begin{center}
\scriptsize
\renewcommand{\arraystretch}{1.8}
\setlength{\tabcolsep}{18pt}
\caption{ The effect of dataset labelling configurations using all FFT-processed data from drive end and fan end}
\label{table:16}
\resizebox{1\textwidth}{!}{
\begin{tabular}{ccccc}
\toprule
\multirow{2}{*}{Data} & \multicolumn{2}{c}{Llama3-8B}                 & \multicolumn{2}{c}{Llama3-instruct}           \\ \cline{2-5} 
                      & Precision       & Recall      & Precision       & Recall      \\
\midrule
CWRUfft-all-10labels  & 0.9910& 0.9910& 0.9944& 0.9944\\
CWRUfft-all-4labels   & 0.99& 0.990& 0.9988& 0.9988\\
\bottomrule
\end{tabular}}
\end{center}
\end{table}

\end{document}